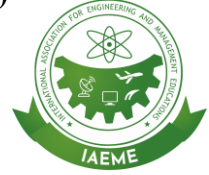

© IAEME Publication

# A THEORETICAL FRAMEWORK FOR AI-DRIVEN DATA QUALITY MONITORING IN HIGH-VOLUME DATA ENVIRONMENTS

[1]Nikhil Bangad, [2]Dr. Vivekananda Jayaram, [3]Manjunatha Sughaturu Krishnappa,
[4]Amey Ram Banarse, [5]Darshan Mohan Bidkar, [6]Akshay Nagpal, [7]Vidyasagar Parlapalli

[1]Meta Inc, Texas, USA, [2]JPMorgan Chase, Texas, USA, [3]Oracle America Inc, California, USA, [4]Yugabyte, California, USA, [5]IEEE Member, Washington, USA, [6]IEEE Member, New Jersey, USA, [7]IEEE Senior Member, Georgia, USA

**ABSTRACT**

*This paper presents a theoretical framework for an AI-driven data quality monitoring system designed to address the challenges of maintaining data quality in high-volume environments. We examine the limitations of traditional methods in managing the scale, velocity, and variety of big data and propose a conceptual approach leveraging advanced machine learning techniques. Our framework outlines a system architecture that incorporates anomaly detection, classification, and predictive analytics for real-time, scalable data quality management. Key components include an intelligent data ingestion layer, adaptive preprocessing mechanisms, context-aware feature extraction, and AI-based quality assessment modules. A continuous learning paradigm is central to our framework, ensuring adaptability to evolving data patterns and quality requirements. We also address implications for scalability, privacy, and integration within existing data ecosystems. While practical results are not provided, it lays a robust theoretical foundation for future research and implementations, advancing data quality management and encouraging the exploration of AI-driven solutions in dynamic environments.*

Nikhil Bangad, Dr. Vivekananda Jayaram, Manjunatha Sughaturu Krishnappa, Amey Ram Banarse, Darshan Mohan Bidkar, 6Akshay Nagpa, Vidyasagar Parlapalli

## 1. INTRODUCTION

In the era of big data, organizations across various sectors are grappling with an unprecedented deluge of information. This exponential growth in data volume, velocity, and variety has revolutionized decision-making processes, offering the potential for deeper insights and more informed strategies [1]. However, the utility of this data hinges critically on its quality. As Redman (2016) aptly noted, "bad data costs the U.S. economy $3.1 trillion a year" [2]. This staggering figure underscores the paramount importance of maintaining high data quality, particularly in high-volume data environments where traditional quality management approaches often fall short.

Data quality, a multifaceted concept, encompasses dimensions such as accuracy, completeness, consistency, timeliness, validity, and uniqueness [3]. In traditional data management systems, ensuring these quality aspects was challenging but manageable through manual audits, rule-based systems, and periodic checks. However, the advent of big data has fundamentally altered the landscape of data quality management. The sheer scale of data, often measured in petabytes or exabytes, renders manual inspection infeasible. The high velocity of data streams, sometimes requiring real-time or near-real-time processing, outpaces traditional batch-oriented quality checks. Moreover, the variety of data types—structured, semi-structured, and unstructured—adds layers of complexity to quality assessment processes [4].

These challenges are further exacerbated in high-volume data systems, where the traditional trade-off between data quality and processing speed becomes increasingly untenable. As Taleb (2013) presciently observed, "The more data you have, the more likely you are to drown in it" [5]. This observation rings particularly true in the context of data quality management, where the risk of overlooking critical quality issues grows proportionally with data volume and complexity.

Existing approaches to data quality management in big data environments have shown promise but also significant limitations. Rule-based systems, while effective for known patterns, struggle to adapt to the dynamic nature of big data [6]. Statistical methods offer scalability but often lack the nuanced understanding required for complex, domain-specific quality issues [7]. Machine learning techniques have demonstrated potential in anomaly detection and pattern recognition, but their application to comprehensive data quality management remains in its infancy [8].

This paper proposes a novel theoretical framework for an AI-driven data quality monitoring system specifically designed for high-volume data environments. Our approach leverages advanced machine learning and artificial intelligence techniques to address the limitations of current methods. We envision a system that not only scales to handle massive data volumes but also adapts to evolving data patterns and quality requirements.

The proposed framework integrates several cutting-edge concepts in AI and data management. At its core is a multi-layered architecture that incorporates intelligent data ingestion, adaptive preprocessing, context-aware feature extraction, and AI-based quality assessment. We draw inspiration from recent advancements in deep learning, particularly the work of LeCun et al. (2015) on deep neural networks [9], to design a system capable of learning complex data quality patterns autonomously.

Our framework also addresses the critical aspect of real-time monitoring, drawing on principles of stream processing and online learning [10]. By continuously updating its models based on incoming data, the proposed system can theoretically maintain high accuracy even in rapidly changing data environments. Furthermore, we explore the potential of federated learning techniques to enable privacy-preserving data quality assessments across distributed data sources [11].

A Theoretical Framework for AI-Driven Data Quality Monitoring in High-Volume Data Environments

While this paper does not present practical implementation results, it provides a comprehensive theoretical foundation for future empirical studies and real-world applications. By carefully considering the challenges and potential solutions in AI-driven data quality monitoring, we aim to stimulate further research and innovation in this critical area.

The remainder of this paper is structured as follows: Section 2 provides a detailed review of related work in the field of data quality management and AI applications in data processing. Section 3 presents our proposed theoretical framework, elaborating on each component and its underlying principles. Section 4 discusses potential implementation considerations, including scalability, privacy, and integration aspects. Section 5 outlines theoretical evaluation metrics for assessing the effectiveness of such a system. Finally, Section 6 concludes with a discussion of future research directions and the potential impact of AI-driven approaches on the field of data quality management.

Through this comprehensive exploration, we seek to contribute to the evolving discourse on data quality in the big data era and provide a robust conceptual foundation for the next generation of data quality monitoring systems.

## 2. BACKGROUND AND RELATED WORK

The intersection of data quality management and artificial intelligence, particularly in the context of high-volume data systems, has been an area of growing research interest. This section provides a comprehensive review of related work, examining both traditional approaches to data quality management and recent advancements in AI-driven techniques.

### 2.1 Traditional Data Quality Management

Traditional approaches to data quality management have primarily focused on rule-based systems, statistical methods, and manual audits. Batini et al. (2009) provided a seminal framework for data quality assessment and improvement, which has been widely adopted in various domains [12]. Their work emphasized the multidimensional nature of data quality, including accuracy, completeness, consistency, and timeliness.

Building on this foundation, Herzog et al. (2007) developed statistical methods for data quality assessment, particularly in the context of official statistics [13]. These methods, while effective for structured data in controlled environments, face significant challenges when applied to high-volume, heterogeneous data systems.

In the realm of data cleaning, which is closely related to quality management, Rahm and Do (2000) proposed a classification of data quality problems and corresponding cleaning approaches [14]. Their work laid the groundwork for many subsequent data cleaning tools and techniques. However, as noted by Abedjan et al. (2016), traditional data cleaning approaches often struggle with the scale and complexity of big data environments [15].

### 2.2 Data Quality in Big Data Environments

The advent of big data has necessitated new approaches to data quality management. Cai and Zhu (2015) were among the first to systematically analyze the challenges of ensuring data quality in big data environments [16]. They highlighted issues such as data volume, acquisition speed, data types, and complex data structures that render traditional quality management techniques inadequate.

Taleb (2013) raised important concerns about the potential for increased errors and misinterpretations as data volume grows [17]. His work underscored the need for more sophisticated, context-aware approaches to data quality assessment in big data scenarios.

Nikhil Bangad, Dr. Vivekananda Jayaram, Manjunatha Sughaturu Krishnappa, Amey Ram Banarse, Darshan Mohan Bidkar, 6Akshay Nagpa, Vidyasagar Parlapalli

Addressing these challenges, Firmani et al. (2016) proposed a framework for measuring data quality in large-scale data systems [18]. Their approach incorporated novel metrics designed to handle the volume and velocity characteristics of big data. However, their framework was primarily focused on structured data and did not fully address the variety aspect of big data.

## 2.3 Machine Learning for Data Quality

The application of machine learning techniques to data quality problems has gained significant traction in recent years. Yakout et al. (2013) developed a system called 'InfoGather' that uses machine learning for data augmentation and cleaning [19]. Their approach demonstrated the potential of ML in enhancing data completeness and accuracy.

In the domain of outlier detection, a crucial aspect of data quality, Liu et al. (2012) provided a comprehensive survey of existing techniques, including several machine learning-based approaches [20]. Their work highlighted the effectiveness of unsupervised learning methods in identifying anomalies in large datasets.

More recently, Rekatsinas et al. (2017) introduced 'HoloClean', a machine learning-based system for holistic error detection and repair [21]. HoloClean represents a significant step towards automated, ML-driven data cleaning. However, its focus on structured data limits its applicability in heterogeneous big data environments.

## 2.4 Deep Learning and Data Quality

The emergence of deep learning has opened new avenues for data quality management. Heidari et al. (2019) proposed a deep learning approach for error detection in textual data [22]. Their method, based on recurrent neural networks, showed promising results in identifying inconsistencies and inaccuracies in unstructured text data.

In the realm of image data quality, Wang and Bovik (2002) developed the structural similarity index (SSIM), which has been widely adopted for image quality assessment [23]. Building on this, Talebi and Milanfar (2018) introduced a convolutional neural network-based approach for no-reference image quality assessment, demonstrating the potential of deep learning in this domain [24].

## 2.5 Real-time Data Quality Monitoring

The need for real-time data quality monitoring in streaming data environments has spawned several research efforts. Artikis et al. (2014) developed a system for complex event recognition that incorporates real-time data quality assessment [25]. Their approach, while groundbreaking, was limited to specific types of event data.

More recently, Psallidas and Wu (2018) proposed 'Smoke', a streaming data quality service that operates at scale [26]. Smoke represents a significant advancement in real-time quality monitoring, though its focus on relational data limits its applicability in heterogeneous big data environments.

## 2.6 Privacy-Preserving Data Quality Assessment

As data privacy concerns have grown, research into privacy-preserving data quality techniques has emerged. Hassan et al. (2019) proposed a framework for privacy-preserving data quality assessment in distributed databases [27]. Their approach uses secure multi-party computation to enable quality checks without revealing sensitive data.

In the context of federated learning, which holds promise for privacy-preserving distributed data analysis, Konečný et al. (2016) introduced techniques for federated optimization [28]. While not specifically focused on data quality, their work lays important groundwork for privacy-preserving distributed data processing.

## 2.7 Research Gaps and Future Directions

Despite these significant advancements, several gaps remain in the field of AI-driven data quality management for high-volume data systems:
- Holistic approaches that address all dimensions of data quality (accuracy, completeness, consistency, timeliness) in a unified framework are lacking.
- Most existing techniques struggle to handle the heterogeneity of big data, particularly when dealing with a mix of structured, semi-structured, and unstructured data.
- Real-time quality assessment for high-velocity data streams remains a challenge, particularly when complex quality dimensions need to be evaluated.
- The integration of domain knowledge with AI-driven techniques for context-aware quality assessment is an area that requires further exploration.
- Privacy-preserving techniques for data quality assessment in distributed, heterogeneous data environments are still in their infancy.

Our proposed theoretical framework aims to address these gaps by leveraging advanced AI techniques in a comprehensive, adaptable system for data quality monitoring in high-volume data environments.

## 3. PROPOSED THEORETICAL FRAMEWORK

This section presents our proposed theoretical framework for an AI-driven data quality monitoring system designed for high-volume data environments. The framework integrates advanced machine learning techniques with domain expertise to provide a comprehensive, adaptive, and scalable approach to data quality management.

### 3.1 Overview of the Framework

Our proposed framework, as illustrated in Fig.1, consists of several interconnected components designed to work in harmony to assess and maintain data quality in real-time, at scale, and across diverse data types. The key components of the framework are: Intelligent Data Ingestion Layer, Adaptive Preprocessing Engine, Context-Aware Feature Extraction, AI-based Quality Assessment Module, Real-time Monitoring and Alerting, and Continuous Learning and Model Adaptation. Additionally, the framework incorporates cross-cutting concerns such as domain knowledge integration, privacy-preserving techniques, and explainable AI to enhance its effectiveness and trustworthiness.

Nikhil Bangad, Dr. Vivekananda Jayaram, Manjunatha Sughaturu Krishnappa, Amey Ram Banarse, Darshan Mohan Bidkar, 6Akshay Nagpa, Vidyasagar Parlapalli

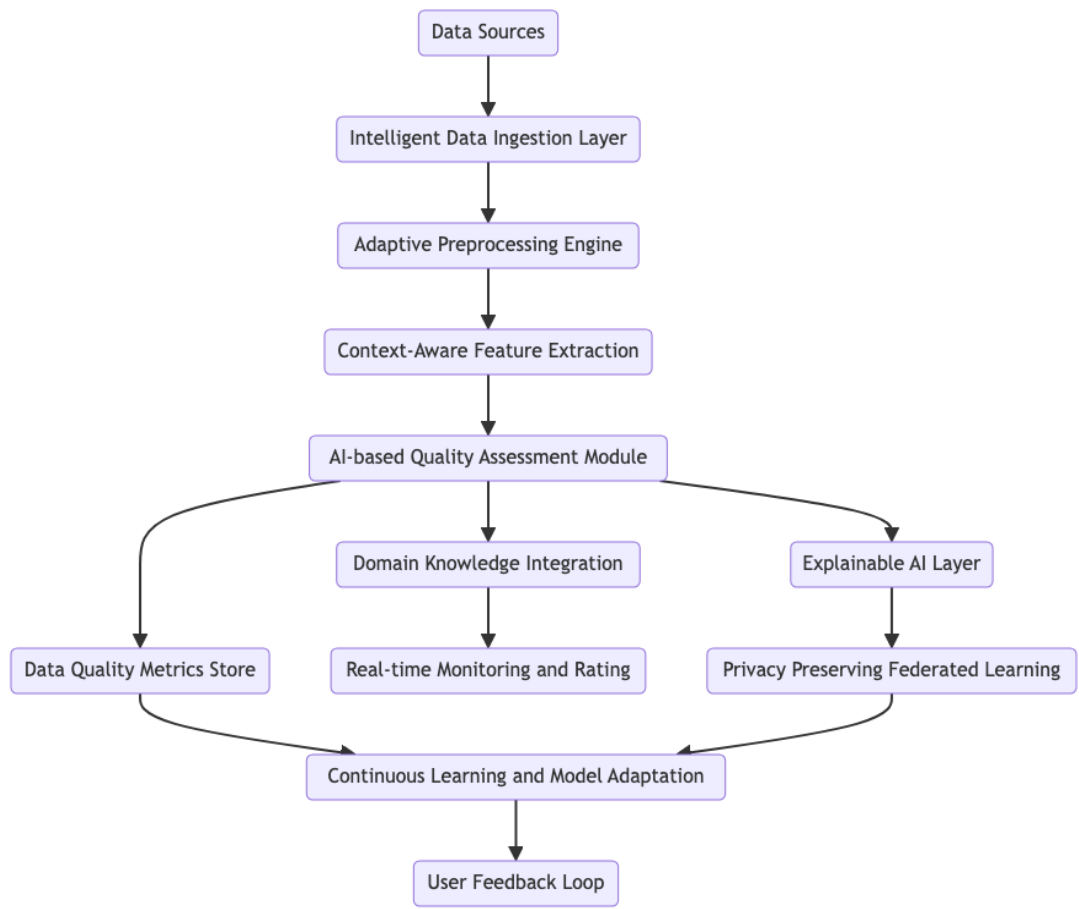

**Figure 1.** AI-Driven Data Quality Monitoring Framework Architecture

### 3.2 Intelligent Data Ingestion Layer

The Intelligent Data Ingestion Layer serves as the entry point for data into our quality monitoring system. This layer is designed to handle the volume and variety characteristics of big data. It utilizes distributed streaming technologies, such as Apache Kafka [29], to manage high-volume data ingestion. To address the variety of big data, this layer employs machine learning techniques for automatic detection and classification of data types and structures, enabling the handling of structured, semi-structured, and unstructured data [30].

To manage the sheer volume of incoming data, the ingestion layer implements adaptive sampling techniques. These ensure that representative samples are selected for quality assessment while maintaining the ability to process high volumes of data [31]. Furthermore, this layer automatically extracts and manages metadata, which is crucial for context-aware quality assessment in subsequent stages of the framework [32].

### 3.3 Adaptive Preprocessing Engine

The Adaptive Preprocessing Engine prepares ingested data for quality assessment, dynamically adjusting its operations based on data characteristics and quality requirements. This engine employs reinforcement learning techniques to learn and apply optimal cleansing strategies for different data types and quality issues [33]. This adaptive approach allows the system to evolve its preprocessing strategies as it encounters new data patterns and quality challenges.

A Theoretical Framework for AI-Driven Data Quality Monitoring in High-Volume Data EnvironmentsFor data normalization, the preprocessing engine utilizes transfer learning techniques. This allows it to apply normalization strategies learned from one dataset to another related but distinct dataset, enhancing the system's ability to handle diverse data sources [34]. To address the common issue of missing data, the engine implements advanced imputation techniques using generative adversarial networks (GANs). This approach allows for more accurate and context-aware filling of missing values [35].

An innovative feature of this preprocessing engine is its anomaly-aware transformation process. By incorporating anomaly detection at this early stage, the system can flag potential quality issues before they propagate through the entire data pipeline [36]. This early warning system enhances the overall efficiency and effectiveness of the quality monitoring process.

## 3.4 Context-Aware Feature Extraction

The Context-Aware Feature Extraction component is crucial for capturing the nuanced aspects of data quality that depend on the specific context and intended use of the data. This component utilizes deep learning techniques such as word embeddings and graph neural networks to capture semantic relationships in the data [37]. By understanding these semantic relationships, the system can make more informed judgments about data quality, particularly for text-based and relational data.

To capture time-dependent quality aspects, this component implements recurrent neural networks (RNNs) and temporal convolutional networks (TCNs) [38]. These architectures are particularly well-suited for identifying quality issues that manifest over time, such as data drift or temporal inconsistencies. For datasets that include multiple types of data (e.g., text, numerical, and categorical), the feature extraction component employs attention mechanisms to combine features from these different modalities [39]. This multi-modal approach ensures that all aspects of the data contribute to the overall quality assessment.

To manage the high dimensionality often associated with big data, this component applies autoencoders for nonlinear dimensionality reduction [40]. This technique allows the system to compress the feature space while preserving the information most relevant to data quality assessment, thereby improving both the efficiency and effectiveness of subsequent processing steps.

## 3.5 AI-based Quality Assessment Module

The AI-based Quality Assessment Module forms the core of our framework, leveraging various AI techniques to assess different dimensions of data quality. This module implements a multi-task learning approach to simultaneously assess multiple quality dimensions, including accuracy, completeness, consistency, and timeliness [41]. By addressing these dimensions concurrently, the system can provide a more holistic view of data quality and identify complex quality issues that span multiple dimensions.

For robust outlier identification, this module combines multiple anomaly detection algorithms, including isolation forests, autoencoders, and DBSCAN, in an ensemble approach [42]. This ensemble method enhances the system's ability to detect a wide range of anomalies across diverse data types and distributions. To address data consistency, the module utilizes siamese networks for learning consistency rules and identifying violations [43]. This approach allows the system to adapt its consistency checks to the specific patterns and relationships present in each dataset.

https://iaeme.com/Home/journal/IJCET 624 editor@iaeme.com

Nikhil Bangad, Dr. Vivekananda Jayaram, Manjunatha Sughaturu Krishnappa, Amey Ram Banarse, Darshan Mohan Bidkar, 6Akshay Nagpa, Vidyasagar Parlapalli

A key innovation in this module is the use of reinforcement learning for adaptive thresholding of quality metrics [44]. This technique allows the system to dynamically adjust quality thresholds based on data characteristics and the potential downstream impact of quality issues. By doing so, the system can prioritize quality issues more effectively and reduce false alarms.

### 3.6 Real-time Monitoring and Alerting

The Real-time Monitoring and Alerting component provide immediate visibility into data quality issues and trends. It leverages stream processing frameworks for continuous computation of quality metrics, enabling real-time insights into the state of data quality [45]. To enhance its proactive capabilities, this component implements time series forecasting models, such as Long Short-Term Memory (LSTM) networks and Prophet, to predict and alert on potential future quality issues [46].

To address the challenge of alert fatigue, the component uses multi-armed bandit algorithms for intelligent alert prioritization [47]. This approach optimizes the generation of alerts, ensuring that users are notified of the most critical issues without being overwhelmed by less significant ones. For intuitive representation of multi-dimensional quality metrics, the component employs advanced visualization techniques [48]. These visualizations aid in the rapid comprehension of complex quality states and trends.

### 3.7 Continuous Learning and Model Adaptation

The Continuous Learning and Model Adaptation component ensures that the system evolves with changing data patterns and quality requirements. It implements online learning algorithms to continuously update models with new data [49], allowing the system to adapt to gradual changes in data distributions or quality standards. For more significant shifts in data characteristics, the component utilizes transfer learning techniques to adapt models to new data sources or dramatically changed data distributions [50].

To efficiently obtain labels for model updating, the component employs active learning strategies [51]. This approach minimizes the need for manual labeling by intelligently selecting the most informative instances for human review. For ongoing optimization of model performance, the component uses genetic algorithms for continuous tuning of model hyperparameters [52]. This evolutionary approach allows the system to maintain peak performance even as data characteristics and quality requirements change over time.

### 3.9 Cross-cutting Concerns

The framework addresses several cross-cutting concerns to enhance its overall effectiveness and applicability. For domain knowledge integration, it incorporates neuro-symbolic AI techniques and knowledge graphs [53,54]. These approaches allow the system to combine the flexibility of machine learning with the precision of domain-specific rules and expertise.

To address privacy concerns, particularly in distributed data environments, the framework implements privacy-preserving federated learning techniques [55]. These allow for collaborative learning across multiple data sources without centralizing sensitive data. Furthermore, differential privacy mechanisms are employed to protect individual data points during model training [56].

Recognizing the importance of transparency in AI systems, particularly for critical applications like data quality management, the framework includes an Explainable AI Layer. This layer employs techniques such as SHAP (SHapley Additive exPlanations) values and attention mechanisms to provide interpretable quality assessments [57,58].

These explanations enhance user trust and facilitate more effective human-AI collaboration in maintaining data quality.

## 3.10 Theoretical Implications and Potential Impact

Our proposed framework represents a significant theoretical advancement in AI-driven data quality management. By integrating cutting-edge AI techniques with domain expertise and privacy-preserving mechanisms, it offers a comprehensive approach to addressing the challenges of data quality in high-volume, heterogeneous data environments.

The framework's adaptive and learning-oriented design suggests the potential for continual improvement in quality assessment accuracy and efficiency. Moreover, its emphasis on explainability and domain knowledge integration addresses critical concerns of trustworthiness and relevance in AI systems.

While practical implementation and empirical validation are beyond the scope of this theoretical proposal, the framework lays a robust foundation for future research and development in AI-driven data quality management. It opens up new avenues for investigation into the synergies between various AI techniques in the context of data quality, and provides a roadmap for developing more intelligent, adaptive, and trustworthy data quality systems.

## 4. IMPLEMENTATION CONSIDERATIONS

While the proposed framework provides a theoretical foundation for AI-driven data quality monitoring in high-volume data systems, its practical implementation presents several challenges and considerations. This section discusses key implementation aspects, focusing on scalability, privacy, and integration with existing data ecosystems.

## 4.1 Scalability Considerations

Implementing an AI-driven data quality monitoring system for high-volume data environments requires careful attention to scalability at every level of the architecture. The system must be capable of handling not only large volumes of data but also high velocity data streams and a variety of data types.

### *4.1.1 Distributed Architecture*

To achieve the necessary scalability, the system should be implemented using a distributed architecture. This approach allows for horizontal scaling, where additional computing resources can be added to the system to handle increased load [59]. A microservices-based architecture can provide the flexibility and scalability required for this system [60].

Fig.2 illustrates a potential distributed architecture for implementing the proposed framework: In this architecture, each component of the framework is implemented as a set of scalable services. The use of a distributed message queue (e.g., Apache Kafka) allows for decoupling of data ingestion from processing, enabling independent scaling of these components [61].

### *4.1.2 Data Partitioning and Sharding*

To handle large volumes of data, effective data partitioning and sharding strategies should be employed. This involves distributing data across multiple nodes based on certain criteria (e.g., time ranges, data categories, or hash functions) [62]. Proper partitioning can significantly improve query performance and allow for parallel processing of data quality assessments.

Nikhil Bangad, Dr. Vivekananda Jayaram, Manjunatha Sughaturu Krishnappa, Amey Ram Banarse, Darshan Mohan Bidkar, 6Akshay Nagpa, Vidyasagar Parlapalli

*4.1.3 In-Memory Processing*

For real-time data quality monitoring, in-memory processing techniques should be considered. Technologies like Apache Spark or Flink can provide the necessary speed for processing high-velocity data streams [63]. These frameworks allow for distributed in-memory computations, significantly reducing latency in quality assessments.

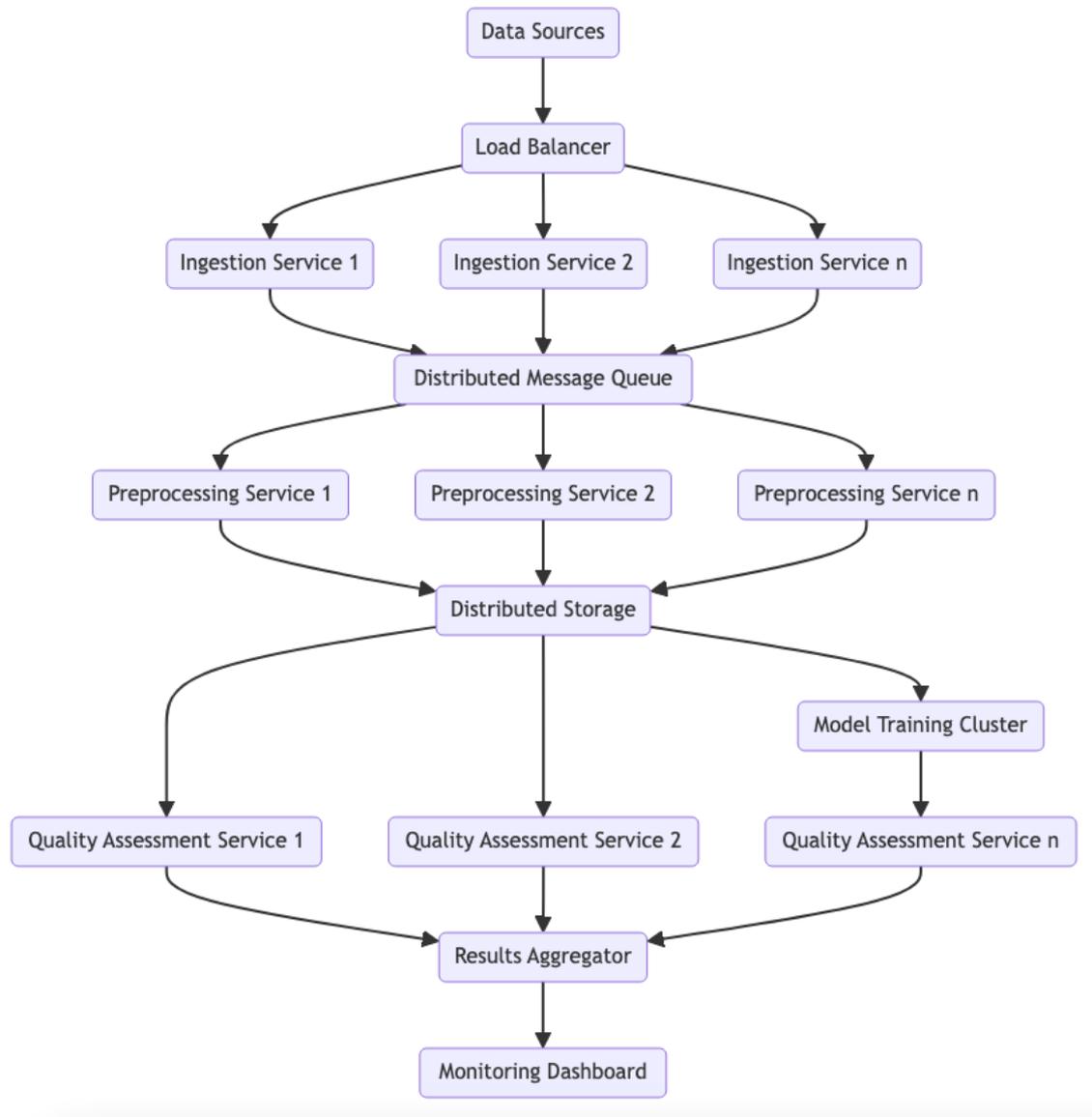

**Figure 2.** Distributed Architecture for Scalable Implementation

*4.1.4 Adaptive Resource Allocation*

To optimize resource utilization, the system should implement adaptive resource allocation mechanisms. This can be achieved through the use of container orchestration platforms like Kubernetes, which allow for dynamic scaling of services based on workload [64].

## 4.2 Privacy Considerations

Given the sensitive nature of data quality information and the potential for the system to process personal or confidential data, privacy considerations are paramount in the implementation of this framework as shown in **Table 1**.

*4.2.1 Data Anonymization and Pseudonymization*

Where possible, data should be anonymized or pseudonymized before quality assessment. This can help protect individual privacy while still allowing for meaningful quality analysis. Techniques such as k-anonymity, l-diversity, and t-closeness should be considered based on the specific requirements and data types [65].

*4.2.2 Differential Privacy*

For scenarios where data aggregation is performed, differential privacy techniques should be implemented to prevent the identification of individuals from aggregate statistics [66]. This is particularly important for the reporting and visualization components of the system.

*4.2.3 Encrypted Processing*

To protect data during processing, homomorphic encryption techniques can be explored. These allow for computations to be performed on encrypted data, though the computational overhead needs to be carefully considered [67].

*4.2.4 Federated Learning*

For distributed environments where data cannot be centralized due to privacy concerns, federated learning approaches should be implemented. This allows the AI models to be trained across multiple decentralized edge devices or servers holding local data samples, without exchanging them [68].

**Table 1.** Comparison of Privacy-Preserving Techniques

| Technique | Pros | Cons | Suitable for |
|---|---|---|---|
| **Data Anonymization** | Simple to implement, Preserves data utility | Potential for re-identification in some cases | Static datasets |
| **Differential Privacy** | Strong privacy guarantees, Suitable for statistical queries | Can reduce data utility, Complex to implement correctly | Aggregate queries, Public datasets |
| **Homomorphic Encryption** | Allows computation on encrypted data, Strong security | High computational overhead, Limited operations | Sensitive data requiring computation |
| **Federated Learning** | Keeps data decentralized, Scalable | Communication overhead, Potential for inference attacks | Distributed data environments |

## 4.3 Integration Considerations

Integrating the AI-driven data quality monitoring system with existing data ecosystems presents several challenges and opportunities.

*4.3.1 API-First Design*

To facilitate integration with a wide range of existing systems, an API-first design approach should be adopted [69]. This involves designing and documenting clear, versioned APIs for all major components of the system. RESTful APIs can provide a standard interface for data ingestion, quality assessment requests, and result retrieval.

Nikhil Bangad, Dr. Vivekananda Jayaram, Manjunatha Sughaturu Krishnappa, Amey Ram Banarse, Darshan Mohan Bidkar, 6Akshay Nagpa, Vidyasagar Parlapalli

*4.3.2 Event-Driven Architecture*

An event-driven architecture can enhance the system's ability to integrate with existing data pipelines [70]. By publishing data quality events to a central event bus, other systems can subscribe to and act upon these events in real-time.

*4.3.3 Metadata Integration*

The system should be designed to integrate with existing metadata management systems. This allows for the enrichment of data quality assessments with business context and lineage information [71]. Standards like the Common Information Model (CIM) can be leveraged to ensure interoperability [72].

*4.3.4 Workflow Integration*

To embed data quality processes into existing data workflows, integration with popular workflow orchestration tools (e.g., Apache Airflow, Luigi) should be considered [73]. This allows for the automation of data quality checks as part of broader data processing pipelines.

*4.3.5 Visualization Integration*

The monitoring and alerting component of the system should be designed to integrate with popular business intelligence and visualization tools. This can be achieved through the implementation of standard connectors or the exposure of data through ODBC/JDBC interfaces [74]. Implementing an AI-driven data quality monitoring system for high-volume data environments presents significant challenges in terms of scalability, privacy, and integration. However, by leveraging distributed architectures, privacy-preserving techniques, and adopting integration-friendly design principles, these challenges can be effectively addressed. The considerations outlined in this section provide a starting point for translating the theoretical framework into a practical, scalable, and privacy-preserving system that can seamlessly integrate with existing data ecosystems.

## 5. THEORETICAL EVALUATION METRICS

Evaluating the effectiveness of an AI-driven data quality monitoring system in high-volume data environments requires a multifaceted approach. This section proposes a framework for theoretical evaluation, encompassing metrics related to data quality assessment accuracy, system performance, and the effectiveness of AI components, building on the foundational work of Wang and Strong [75] in defining data quality dimensions.

## 5.1 Evaluation Framework Overview

The proposed evaluation framework consists of three main categories of metrics:
1. Data Quality Assessment Metrics
2. System Performance Metrics
3. AI Effectiveness Metrics

Fig.3 illustrates the relationship between these categories and their subcategories.

A Theoretical Framework for AI-Driven Data Quality Monitoring in High-Volume Data Environments

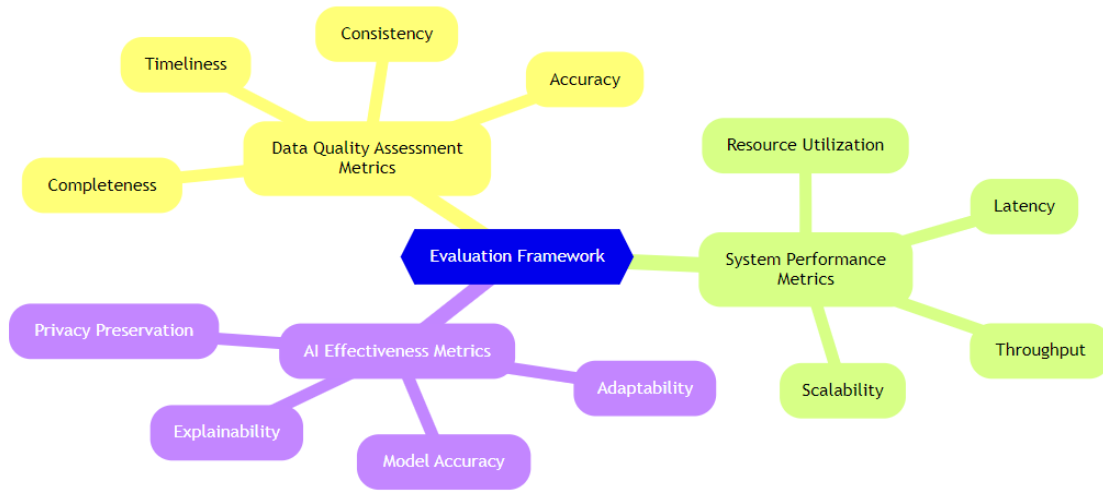

**Figure 3.** Evaluation Framework

## 5.2 Evaluation Framework Overview

Table 2 summarizes the key metrics for each category, drawing on various aspects of data quality assessment and system evaluation as discussed in the literature [76,77,78]

**Table 2.** Summary of Evaluation Metrics

| Category | Metric | Description |
| --- | --- | --- |
| Data Quality Assessment | F1 Score | Measures the accuracy of data quality issue detection |
| Data Quality Assessment | Completeness Detection Rate (CDR) | Assesses the system's ability to identify missing data |
| Data Quality Assessment | Consistency Rule Compliance Rate (CRCR) | Evaluates the detection of data inconsistencies |
| Data Quality Assessment | Timeliness Detection Accuracy (TDA) | Measures the accuracy in identifying outdated data |
| System Performance | Scale-Up Efficiency (SUE) | Assesses the system's ability to handle increasing data volumes |
| System Performance | End-to-End Latency (E2EL) | Measures the time delay in data quality assessment |
| System Performance | Records Processed Per Second (RPPS) | Evaluates the system's throughput |
| System Performance | Resource Efficiency Index (REI) | Measures the efficiency of resource utilization |
| AI Effectiveness | Area Under the ROC Curve (AUC-ROC) | Assesses the overall accuracy of AI models |
| AI Effectiveness | Drift Adaptation Score (DAS) | Evaluates the system's ability to adapt to changing data patterns |
| AI Effectiveness | Explainability Score (ES) | Measures the interpretability of AI-driven quality assessments |
| AI Effectiveness | Privacy Loss | Assesses the effectiveness of privacy-preserving techniques |

Nikhil Bangad, Dr. Vivekananda Jayaram, Manjunatha Sughaturu Krishnappa, Amey Ram Banarse, Darshan Mohan Bidkar, 6Akshay Nagpa, Vidyasagar Parlapalli

## 5.3 Evaluation Challenges and Limitations

While these metrics provide a framework for evaluation, several challenges should be acknowledged, as highlighted by various researchers in the field [76,79]

1. Defining absolute data quality in diverse, high-volume environments is complex [76]
2. Some metrics may conflict, requiring careful balance (e.g., privacy vs. accuracy) [78]
3. The importance of metrics may vary based on specific use cases or domains [75,79]
4. Evaluating adaptability and long-term performance requires extended studies [77]
5. Establishing fair comparisons with traditional systems can be challenging due to fundamental differences in approach [79]

## 6. CONCLUSION

This paper has presented a theoretical framework for AI-driven data quality monitoring in high-volume data environments. As organizations increasingly rely on big data for critical decision-making, ensuring data quality becomes paramount [80]. Traditional approaches to data quality management often struggle with the volume, velocity, and variety of modern data ecosystems [81]. Our proposed framework leverages advanced AI techniques to address these challenges, offering a scalable, adaptive, and context-aware approach to data quality monitoring.

Key contributions of this work include:

1. A comprehensive architecture for real-time, AI-driven data quality assessment

2. Integration of multiple AI techniques for holistic quality evaluation

3. Adaptive mechanisms to handle evolving data patterns and quality requirements

4. Consideration of privacy-preserving methods for sensitive data environments

5. A theoretical evaluation framework for assessing the effectiveness of such systems

The potential impact of AI-driven approaches on data quality management is significant. By enabling proactive quality management and contextual quality assessments, these systems could transform how organizations maintain and improve their data assets [82]. The ability to process and evaluate vast amounts of data in real-time opens new possibilities for ensuring data quality at scale [83].

However, realizing this potential will require addressing several challenges. These include ensuring robust data privacy and security [84], building trust in AI-driven quality assessments [85], and bridging the skill gap in data management professionals. Moreover, ethical considerations in AI deployment for data quality management will need careful attention [86].

Looking ahead, promising areas for future research include:

1. Exploring advanced AI techniques such as deep reinforcement learning [87] and neuro-symbolic AI [88] for more sophisticated quality management strategies

2. Developing methods for cross-domain knowledge transfer to enhance system adaptability [89]

3. Advancing explainable AI techniques to improve the interpretability and trustworthiness of quality assessments [90]

In conclusion, the intersection of AI and data quality management represents a frontier of innovation with the potential to significantly enhance the reliability and value of organizational data assets. While challenges remain, the benefits of more accurate, scalable, and adaptive data quality management are compelling. As research progresses and technologies mature, AI-driven approaches are poised to become an integral part of the data quality landscape, enabling organizations to harness the full potential of their data in an increasingly data-driven world.

A Theoretical Framework for AI-Driven Data Quality Monitoring in High-Volume Data Environments

Nikhil Bangad, Dr. Vivekananda Jayaram, Manjunatha Sughaturu Krishnappa, Amey Ram Banarse, Darshan Mohan Bidkar, 6Akshay Nagpa, Vidyasagar Parlapalli

Nikhil Bangad, Dr. Vivekananda Jayaram, Manjunatha Sughaturu Krishnappa, Amey Ram Banarse, Darshan Mohan Bidkar, 6Akshay Nagpa, Vidyasagar Parlapalli

# A Theoretical Framework for AI-Driven Data Quality Monitoring in High-Volume Data Environments